\documentclass{article}
\usepackage{spconf,amsmath,graphicx,hyperref}

\title{GATA2Floor: Graph Attention for Floor Counting in Street-View Facades}
\name{Ngoc Tan Le, Tzoulio Chamiti, Eirini Papagiannopoulou, Nikos Deligiannis\thanks{The work is supported by the ”Onderzoeksprogramma Artificiele Intelligentie (AI) Vlaanderen” programme and by the ERC CoG IONIAN (No. 101171240). Funded by the European Union. Views and opinions expressed are however those of the author(s) only and do not necessarily reflect those of the European Union or the European Research Council Executive Agency. Neither the European Union nor the granting authority can be held responsible for them. Accepted for publication in IEEE ICIP 2026. © 2026 IEEE. Personal use is permitted.}}
\address{ETRO Department, Vrije Universiteit Brussel (VUB), Pleinlaan 2, B-1050 Brussels, Belgium \\ imec, Kapeldreef 75, B-3001 Leuven, Belgium}

\begin{document}
%
\maketitle
\begin{abstract}
Automated analysis of building facades from street-level imagery has great potential for urban analytics, energy assessment, and emergency planning. However, it requires reasoning over spatially arranged elements rather than solely isolated detections. In this work, we model each facade as a graph over window/door detections with a vertical prior on edges. Additionally, we introduce GATA2Floor, a multi-head Graph Attention v2 (GATv2) based model that predicts the global floor count of a building and, via learnable cross-attention queries, softly assigns elements to latent floor slots, yielding interpretable outputs and robustness to irregular designs. To mitigate the lack of labeled datasets, we demonstrate that the proposed graph-based reasoning can be applied without annotations by leveraging a lightweight label-free proposal mechanism based on self-supervised features and vision–language scoring. Our approach demonstrates the value of graph-attention-based relational reasoning for facade understanding.
\end{abstract}
\begin{keywords}
Street view imagery, facade analysis, floor counting, graph attention networks, label-free proposals
\end{keywords}

\section{Introduction}
\label{sec:intro}
Street view imagery (SVI) offers a valuable resource in building facades with multiple potential applications (energy estimation, construction cost/style prediction etc.), where accurate building-level information is critical. Estimating floors, however, requires reasoning over spatially arranged elements (windows/doors) rather than treating detections in isolation. 

Conventional clustering or heuristic-based methods degrade under viewpoint changes, occlusions and irregular layouts, while recent object detection methods (e.g., Faster/Mask R-CNN, YOLO) excel at local facade element modeling but remain fully supervised, lacking robustness in unlabeled datasets, threby leaving the problem of global floor counting and per-element floor assignment unresolved~\cite{ren2015fasterrcnn, he2017maskrcnn, redmon2016yolo, Sezen2022facadecomparision}. 
Recently, Vision–Language Models (VLMs) have demonstrated the ability to identify facade elements in a zero-shot manner~\cite{Pan2024zeroshot}, but their weak spatial grounding limits dense localization. As a result, they are better suited for coarse proposal filtering or verification rather than serving as primary detectors for structured facade analysis. In parallel, graph neural networks (GNNs)~\cite{Scarselli2009gnn} and attention-based variants (GAT and GATv2)~\cite{velivckovic2017graph, brody2021gatv2} propagate information through nodes and edges while adaptively weighting neighbor influence, which is crucial for noisy, irregular neighborhoods typical for a building facade. Notably, the floor counting task receives relatively little attention with existing methods~\cite{Wu_2021, Moubayed28022025, li2023semisupervisedlearningstreetviewimages, sun2025building} often relying on heuristic priors, clustering or monolithic regressors, and rarely delivering interpretable, floor-wise groupings.

This paper models each facade as a vertical-aware graph over window and door detections and presents three contributions: First, we propose GATA2Floor, a GATv2-based model that applies self-attention with a vertical bias mask and a cross-attention module using learnable floor queries. The self-attention aggregates relational information to infer the global floor count, while cross-attention handles soft window- and door-to-floor assignments, improving robustness on irregular facades. To the best of our knowledge, this is the first graph-based formulation targeting \textit{both} floor counting and soft window- and door-to-floor assignment on facades (as opposed to existing floor counting approaches that only output the number of floors of a building). Second, we perform extensive evaluation on the Amsterdam Facade, ECP, eTRIMS, and ParisArtDecoFacades datasets, showing robustness to viewpoint changes, occlusions, and irregular layouts, and consistent gains over clustering baselines~\cite{korc-forstner-tr09-etrims, gadde2016learning}. Third, we show that the proposed graph-based formulation remains applicable in the absence of annotations by relying on a lightweight label-free proposal mechanism, that provides coarse window/door candidates to construct the graph. 

The remainder of the paper is organized as follows: Section~\ref{sec:methodology} discusses the proposed approach, Section~\ref{sec:experiments} presents experimental results and Section~\ref{sec:conclusion} concludes the work.

\section{Proposed Methodology}
\label{sec:methodology}

\begin{figure*}[t]
    \centering
    \includegraphics[width=1\linewidth]{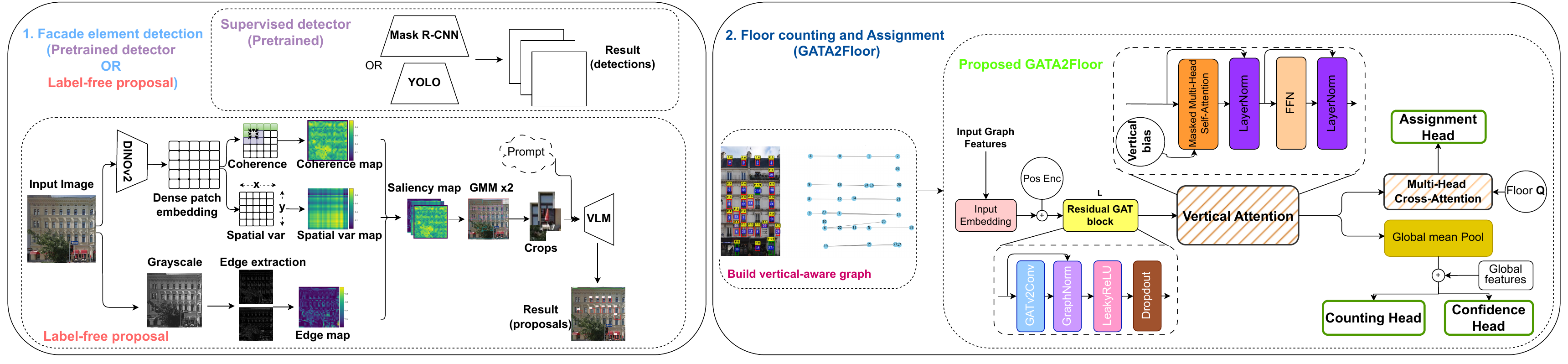}
    \caption{Overview of the framework, it consists of (1) Facade element proposal module using either a supervised detector (default) or a lightweight label-free proposal mechanism (fallback for unlabeled data, see Section~\ref{subsec:zeroshot}) that produces computed detections/proposals. (2) a vertical-aware graph representation is constructed (Sections~\ref{subsec:graph-preliminaries},~\ref{subsec:vertical-aware-graph}), and used by the \textbf{GATA2Floor} model with a relative \emph{vertical bias} as a mask in self-attention logits (Vertical Attention). Finally, the model outputs the floor-count, per floor assignment of each detection and its confidence.}
    \label{fig:overview}
\end{figure*}

The proposed GATA2Floor operates on precomputed window and door bounding boxes obtained either from a supervised detector or from a lightweight label-free proposal mechanism (Section~\ref{subsec:zeroshot}) when annotations are unavailable, and builds a graph over those boxes rather than on the raw image. Concretely, given a set of $N$ element detections (a.k.a., proposals) $B=\{b_{i}\}_{i=1}^{N}$ with $b_{i}=(x_{i}^{\min}, y_{i}^{\min}, w_i, h_i)$ in an image of size $W \times H$, this work constructs a vertical-aware graph representation over $B$, that is fed to the GATA2Floor model to jointly predict the global floor count and soft per-element floor memberships (see Fig.~\ref{fig:overview}).

\subsection{Problem formulation and Graph preliminaries}
\label{subsec:graph-preliminaries}
Let $G = (V,E)$ be a graph with $|V| = N$ nodes and edges~$E$. Each node $v_i \in V$ encodes element $b_i$ via a feature vector~$\mathbf{f}_i \in \mathbf{R}^6$:
\[
\mathbf{f}_i = [n_c^x,n_c^y,n_w,n_h,ar,is_\text{window}],
\]
where $(x_i^c, y_i^c) = (x_i^{\min} + \frac{w_i}{2}, y_i^{\min} + \frac{h_i}{2})$ are box centers, $n^x_c = \frac{x_i}{W}, n^y_c = \frac{y_i}{H}$ are normalized center coordinates, $n_w = \frac{w_i}{W}, n_h = \frac{h_i}{H}$ are normalized dimensions, $ar= \frac{w_i}{h_i}$ is the aspect ratio, and $is_\text{window} \in \{0,1\}$ indicates window (1) or door (0).
For pairs of nodes $(v_i, v_j)$, edge features are:
\[
\mathbf{e}_{ij} = [d^x_\text{norm}, d^y_\text{norm}, \text{IoU}(b_i, b_j), v_\text{Overlap}(b_i, b_j)] \in \mathbf{R}^4,
\]
with $d^x_\text{norm} = \frac{|x^c_i - x^c_j|}{W}$, $d^y_\text{norm} = \frac{|y^c_i - y^c_j|}{H}$,  $\text{IoU}(b_i, b_j) = \frac{|b_i \cap b_j|}{|b_i \cup b_j|}$, and $v_\text{Overlap}(b_i, b_j) = \frac{|b_i \cap b_j|}{\min(|b_i|,|b_j|)}$ where $|b_i| = w_i \times h_i$ is the pixel area of box $b_i$. These features prioritize vertical spacing and overlap for floor memberships, while horizontal distance and IoU specify columns and co-linear facade patterns.

\subsection{Vertical-aware graph construction}
\label{subsec:vertical-aware-graph}
This paper encodes directional geometry relationships, which depend only on pair-wise vertical gaps, into edges to build a graph. The edge construction rule is formed by exploiting the architectural \emph{prior} that elements on the same floor are closely aligned along $y$. An edge $e_{ij}$ is established if  $\mathbf{A}_{ij}$ is satisfied, 
\begin{equation}
\mathbf{A}_{ij} = 1\{d^y_\text{norm} \le \tau_\text{vertical}\},
\label{eq:connectivity}
\end{equation}
where $\tau_\text{vertical}$ is computed adaptively per facade from the empirical distribution of $\{d^y_\text{norm}\}$ to enforce within-floor locality:
\begin{equation}
\tau_\text{vertical} = \alpha_\text{outlier} \times \mu_{\text{top-}k},
\label{eq:vertical_tau}
\end{equation}
with $\mu_{\text{top-}k}$ the mean of the top-$k$ vertical gaps (e.g., $k =3$) and $\alpha_\text{outlier} \in [0, 1]$ an outlier to mitigate missing‑floor outliers; this adapts the prior to each building’s internal pitch and preserves within-floor connectivity.

We also associate each facade with a global vector $g$ to capture long-range context beyond local evidence. It stores the normalized vertical gap information $\tau_\text{vertical}$ for edge construction [see~\eqref{eq:vertical_tau}]; element density $\rho$ for scale information and vertical dispersion $\sigma_y = \text{std}(\{\frac{y_i}{H}\}^N_{i=1})$ for vertical spread:
\[
g = [\tau_\text{vertical}, \rho, \sigma_y] \in \mathbf{R}^3.
\]

Pseudo labels provide weak supervision derived from vertical connectivity. Let $C = \{C_s\}^M_{s=1}$ be the connected components of this graph, then define:
\begin{equation}
    \bar{c} = M \text{ and } \bar{y}_i =s \iff v_i \in C_s.
    \label{eq:pseudo}
\end{equation}
The pseudo floor count $\bar{c}$ serves as weak supervision for the counting head, while the pseudo floor ID $\bar{y}_i$ supervises the assignment head (Section~\ref{subsec:pred_heads}) during the training phase only.

This graph is then fed to the GATA2Floor model (see Fig.\ref{fig:overview}), allowing attention to learn vertical priors under irregular layouts, while the distribution of vertical distances $\{d^y_\text{norm}\}$ is later used to compute a bias map inside the model.

\subsection{Local‑to‑global encoder}
\label{subsec:enc_blocks}
\textbf{Residual GATv2}:
\label{paragraph:gatv2}
Given the node embeddings $H^{(0)} = [h_i^{(0)}]_{i=1}^N$ encoded by summing the \emph{input embedding}---a learned projection---of the raw node features with a positional MLP on $[n_c^x,n_c^y]$, the model applies $L$ times residual, edge-aware GATv2 blocks to capture local relationships within connected components under the connectivity rule defined by $\mathbf{A}$ \eqref{eq:connectivity}. Each block computes local attention over a neighbor set $\mathcal{N}(i) = \{ j: A_{ij} = 1\}$ then follows with a skip connection and normalization to stabilize gradients and emphasize intra‑component influence as in Fig.~\ref{fig:overview} to produce $H^{(L)} = [h_i^{(L)}]_{i=1}^N$ where: 
\begin{equation}
  h_{i}^{(l+1)} = \text{GraphNorm}(h_{i}^{(l)} + \text{GATv2}(h^{(l)}_{i}, h_{j}^{(l)}, e_{ij})); j \in \mathcal{N}(i)
\end{equation}

\noindent
\textbf{Vertical Attention}:
\label{paragraph:vertical_attn}
To further aggregate long‑range vertical regularities across components, a masked multi‑head self‑attention layer is then applied to the block output $H^{(L)}$, with a key \emph{vertical bias} $b_{ij}$ computed from the pair-wise vertical distance $d^y_\text{norm}(i,j) = \frac{|y_i - y_j|}{H}$. For each head $h$, a query/key/value ($\mathbf{Q, K, V})$ projection is formed with a bias as mask logits:
\begin{equation}
\alpha_{ij}^{(h)} = \text{softmax}(\frac{(q_i^{(h)})^{\top}(k_j^{(h)})}{(\sqrt{d_h})} + b_{ij}),
\end{equation}
\begin{equation}
b_{ij} = -r(d^y_\text{norm}(i,j)),
\end{equation}
where $r$ is a small MLP so that the \emph{vertical bias} $b_{ij}$ assigns higher attention to closer node pairs while still allowing cross-floor interactions, resulting in $H^{\mathrm{vert}}$ embeddings that capture global vertical structure and feed \emph{three} complementary heads.

\subsection{Prediction heads}
\label{subsec:pred_heads}
Given the output embeddings $H^\text{vert}$ from \emph{Vertical Attention}, the model employs three heads: a cross-attention head for floor-slot assignment, and a global mean pooling feeding the two heads for the global floor count and counting confidence.

\noindent
\textbf{Counting head and Confidence head}: Global mean pooling over $H^\text{vert}$ produces a permutation-invariant embedding $z$, concatenated with the global vector $g$ (see~\ref{subsec:vertical-aware-graph}) to regress the floor count $\hat{c}$ and its reliability $\hat{u}$ for risk-aware deployment:
\[
\hat{c} = g_\theta ([z; g]) \text{ and } \hat{u} = \sigma(\mathbf{w}_u^{\top}[z; g] + b_u).
\]
This model employs the SmoothL1 loss for floor counting and the MSE for confidence, with the pseudo floor count $\bar{c}$ in~\eqref{eq:pseudo}:
\begin{equation}
    \mathcal{L}_{count} = \text{SmoothL1}_\delta(\hat{c} - \bar{c}),
\end{equation}
\begin{equation}
    \mathcal{L}_{conf} = \Vert \hat{u} - \text{exp(}\vert \hat{c} - \bar{c}\vert)\Vert^2_2.
\end{equation}
with $\text{SmoothL1}_\delta(x) = \frac{x^2}{2\delta}$ if $|x|<\delta$, else $|x|-\frac{\delta}{2}$, $\delta=1$~\cite{girshick2015fast}.

\noindent
\textbf{Assignment head}: Introduces a learnable floor query $\mathbf{Q}_\text{floor}$ with predefined $\mathbf{S}$ slots and computes cross-attention over $\mathbf{K, V}$ sets derived from $H^\text{vert}$, resulting in assignments that align nodes with latent floor slots. We use a CrossEntropy loss with a predicted floor ID $\hat{y}$ and pseudo value $\bar{y}$ [see~\eqref{eq:pseudo}]:
\begin{equation}
    \mathcal{L}_\text{assign} = \text{CE}(\hat{y}, \bar{y}).
\end{equation}

\noindent Finally, the three heads are trained end‑to‑end by minimizing:
\begin{equation}
    \mathcal{L} = \omega_\text{count}\mathcal{L}_\text{count} + \omega_\text{assign}\mathcal{L}_\text{assign} + \omega_\text{conf}\mathcal{L}_\text{conf},
\end{equation}
where each head is given a corresponding weight $\omega$ to balance
gradient scales across tasks.

\subsection{Label-free facade element proposal}
\label{subsec:zeroshot}
Motivated by the lack of labeled data, a lightweight label-free proposal pipeline is designed to generate coarse window/door candidates from unlabeled SVI. We fuse three complementary cues into a saliency map: (i) a Sobel edge map emphasizing rectilinear boundaries typical of architectural openings~\cite{sobel1973operator}; (ii) a spatial-variance map highlighting grid-like regularities; and (iii) a DINOv2 coherence map aggregating local context via patch-embedding cosine similarity\cite{oquab2023dinov2}. Then a hierarchical Gaussian Mixture Model (GMM) mitigates noise: a global GMM ($n=2$) isolates the facade region and a local GMM segments window/door-like blobs~\cite{dempster1977em}. Cropped proposals are scored by CLIP and GPT using positive (window/door) and negative (balcony/sign) prompts~\cite{radford2021CLIP, openai2024gpt4ocard}, and the highest-mean score cluster defines the final proposals fed to GATA2Floor, enabling label-free graph construction.

\section{Experiments and results}
\label{sec:experiments}
\subsection{Datasets}
We use multiple common labeled datasets in the facade detection field like the Amsterdam Facade, ECP, eTRIMS, and ParisArtDecoFacades~\cite{korc-forstner-tr09-etrims, gadde2016learning}. We perform manual labeling for the floor-level ground truth generation.

\begin{figure}[t]
    \centering
    \includegraphics[width=1\linewidth]{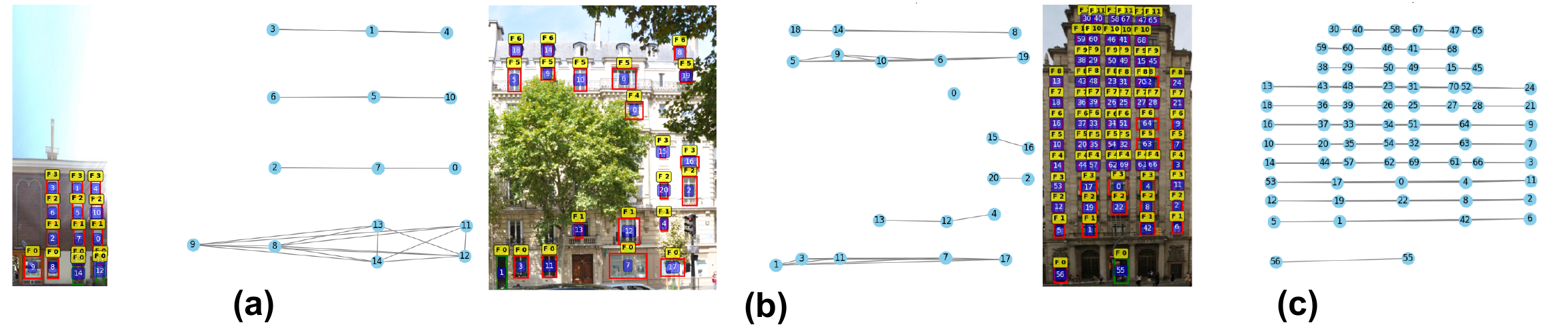}
    \caption{The proposed building-as-graph representation that recovers accurate pseudo counts: 4 for (a) irregular facade structures, 7 for (b) occlusions, 12 for (c) complex elements.}
    \label{fig:graph_fig}
\end{figure}

\subsection{Graph-based representation}
We first evaluate the proposed graph-based representation before the end-to-end GATA2Floor model. As shown in Fig.~\ref{fig:graph_fig}, this representation is robust across common real-world scenarios provided at least one window/door element is detected per true floor. Under this assumption, vertical connectivity induces floor-wise groupings even in challenging cases. Failure cases arise only when an entire floor has no detected elements, breaking the integrity condition and fragmenting the vertical connectivity needed to infer a correct component structure.

\subsection{Floor Counting: Baseline Comparison}
\label{subsec:floor_counting}
The GATA2Floor backbone employs $L=3$ residual GATv2 blocks ($H_\text{GAT}=8$ heads each) with light edge dropout $p=0.1$, followed by the Vertical Attention block ($H_\text{vert}=8$); the cross-attention assignment head uses $S=15$ learnable floor queries $\mathbf{Q}_\text{floor}$for soft floor memberships, with $S$ chosen as a dataset-agnostic upper bound covering the floor-count range across all considered datasets. Training adopts batch size 4, gradient clipping 2.0, AdamW ($lr = 3 \times 10^{-5}$, weight decay $1 \times 10^{-4}$), runs for 200 epochs with loss weights $\omega_\text{count}=0.4$, $\omega_\text{assign}=0.4$, $\omega_\text{conf}=0.2$ to balance training objectives~\cite{loshchilov2019decoupledweightdecayregularization}.

\begin{figure}[t]
    \centering
    \includegraphics[width=1\linewidth]{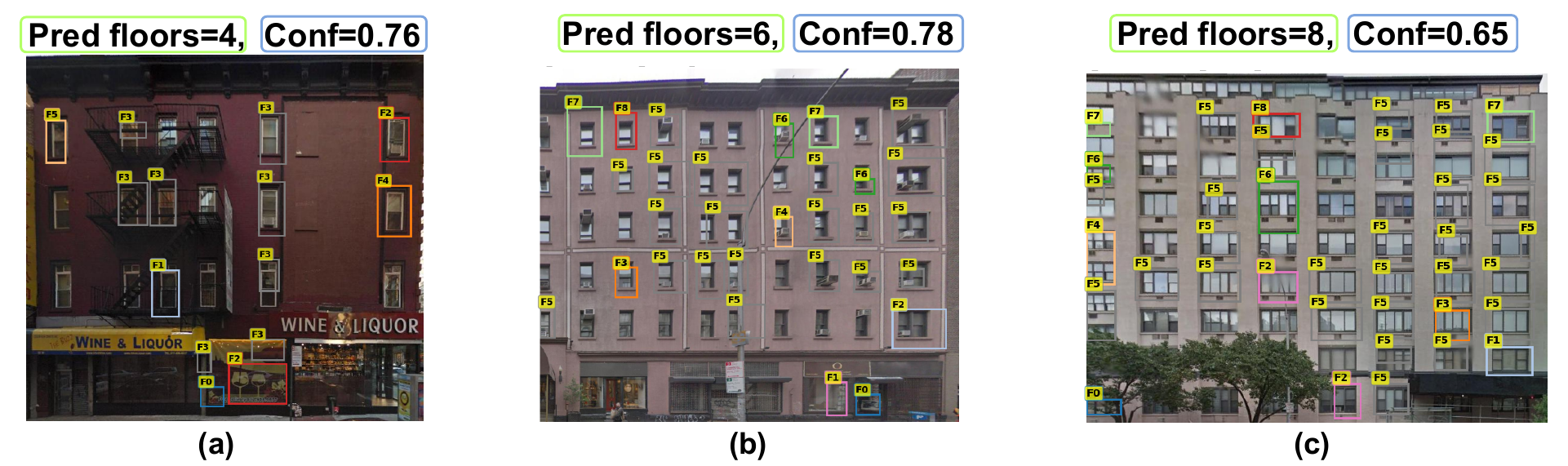}
    \caption{GATA2Floor visualization results, with multi-task objectives: Estimated floor number (green), counting confidence (blue) and per-element floor assignment (yellow boxes)}
    \label{fig:gata2floor_img}
\end{figure}

We compare GATA2Floor against a ResNet50 classifier (a CNN baseline) trained with CrossEntropy on manually annotated floor counts and clustering methods across labeled datasets, reporting three metrics (Tab.~\ref{tab:comparison_floor_count} and Fig.~\ref{fig:gata2floor_img}): (i)~\textbf{Mean Absolute Error (MAE)}: average absolute difference between predicted and ground-truth targets (lower is better); (ii)~\textbf{F1-Score}: harmonic mean of precision and recall ($F1 = 2 \cdot \frac{\text{precision} \cdot \text{recall}}{\text{precision} + \text{recall}}$), treating each floor count as a separate class (higher is better); (iii)~\textbf{Accuracy}: percentage of buildings with exactly correct floor count predictions (higher is better).

GATA2Floor reaches 86\% accuracy (MAE 0.14) on Amsterdam dataset, rising to 90\% (MAE 0.26) on ECP with higher floor counts, decreasing to 58\% (MAE 0.42) on ParisADF dataset due to architectural occlusions, and 56\% on eTRIMS where unrectified viewpoints undermine vertical regularity. The ResNet50's high accuracy on ECP reflects a dataset artifact---ECP heavily consists of 6–7 floor buildings, inflating accuracy via majority-class prediction without structural reasoning---whereas GATA2Floor generalizes more robustly across diverse floor distributions and irregular layouts. It performs best on facades with clear, regular structure, while under unrectified viewpoints, as in eTRIMS, and heavy occlusions, as in ParisADF, the performance is reduced. 

\begin{table}[t]
\centering
\footnotesize
\caption{Floor count results between GATA2Floor, ResNet50 baseline, clustering methods: Kernel-Density Estimation (KDE) ~\cite{Dobson2023floor}, Agglomerative Clustering (AC), Intersection Clustering (IC). Best metric in bold; second-best underlined.}
\begin{tabular}{l|ccccc}
\hline
\textbf{Dataset} & \textbf{GATA2Floor} & \textbf{ResNet50} & \textbf{KDE} & \textbf{AC} & \textbf{IC}\\
\hline
\multicolumn{6}{c}{\textbf{Mean Absolute Error (MAE)} $[0, \infty) \downarrow$} \\
\hline
AmsterdamF & \textbf{0.14} & 0.22 & \underline{0.17} & 0.22 & 0.20 \\
ECP & \underline{0.26} & \textbf{0.05} & 0.80 & 0.41 & 0.30 \\
eTRIMS & \underline{0.42} & 0.83 & 0.65 & 0.52 & \textbf{0.28} \\
ParisADF & \textbf{0.42} & 0.88 & - & 0.84 & \underline{0.78}\\
\hline
\multicolumn{6}{c}{\textbf{F1-Score} $[0, 1] \uparrow$} \\
\hline
AmsterdamF & \textbf{0.84} & 0.74 & \underline{0.83} & 0.80 & 0.81\\
ECP & 0.65 & 0.49 & 0.49 & \underline{0.73} & \textbf{0.80}\\
eTRIMS & 0.41 & 0.36 & 0.49 & \underline{0.63} & \textbf{0.72} \\
ParisADF & \textbf{0.63} & 0.46 & - & \underline{0.48} & \underline{0.48}\\
\hline
\multicolumn{6}{c}{\textbf{Accuracy} $[0, 1] \uparrow$} \\
\hline
AmsterdamF & \textbf{0.86} & 0.80 & \underline{0.83} & 0.80 & 0.81\\
ECP & \underline{0.90} & \textbf{0.95} & 0.38 & 0.63 & 0.73\\
eTRIMS & 0.58 & 0.42 & 0.48 & \underline{0.62} & \textbf{0.72}\\
ParisADF & \underline{0.56} & \textbf{0.61} & - & 0.44 & 0.44\\
\hline
\end{tabular}%
\label{tab:comparison_floor_count}
\end{table}

\subsection{Component-wise Analysis}
\label{sec:ablation}

We analyze the contribution of each architectural component in GATA2Floor using three variants trained on Amsterdam, ECP, eTRIMS, and ParisArtDecoFacades datasets: (i) \textbf{v1: GATv2}, which aggregates information only within connected components; (ii) \textbf{v2: GATv2 + Vertical Attention}, which introduces global vertical interactions; and (iii) \textbf{v3: Full GATA2Floor}, which additionally includes the floor-query assignment head. All variants share the same training protocol (Tab.~\ref{tab:ablation_components}).

\begin{table}[t]
\centering
\footnotesize
\caption{Component-wise analysis on three datasets. Each variant adds one architectural component incrementally. Best metric in bold; second-best underlined.}
\label{tab:ablation_components}
\begin{tabular}{l|ccc}
\hline
\textbf{Dataset} & \textbf{v1: GATv2} & \textbf{v2: +Vert.Attn} & \textbf{v3: GATA2Floor} \\
\hline
\multicolumn{4}{c}{\textbf{Mean Absolute Error (MAE)} $[0, \infty) \downarrow$} \\
\hline
AmsterdamF & 0.22 & \underline{0.29} & \textbf{0.14} \\
ECP & 0.28 & \underline{0.27} & \textbf{0.26} \\
eTRIMS & 0.45 & \textbf{0.34} & \underline{0.42} \\
ParisADF & \underline{0.48} & 0.50 & \textbf{0.42} \\
\hline
\multicolumn{4}{c}{\textbf{F1-Score} $[0, 1] \uparrow$} \\
\hline
AmsterdamF & \underline{0.71} & 0.55 & \textbf{0.84} \\
ECP & 0.42 & \underline{0.57} & \textbf{0.65} \\
eTRIMS & 0.37 & \textbf{0.70} & \underline{0.41} \\
ParisADF & 0.36 & \underline{0.41} & \textbf{0.63} \\
\hline
\multicolumn{4}{c}{\textbf{Accuracy} $[0, 1] \uparrow$} \\
\hline
AmsterdamF & \underline{0.75} & 0.70 & \textbf{0.86} \\
ECP & 0.80 & \underline{0.85} & \textbf{0.90} \\
eTRIMS & \underline{0.58} & \textbf{0.63} & \underline{0.58} \\
ParisADF & \underline{0.48} & 0.44 & \textbf{0.56} \\
\hline
\end{tabular}
\end{table}

As shown in Tab.~\ref{tab:ablation_components}, vertical attention (v2) consistently improves v1 by enabling long-range vertical interactions, particularly when disconnected components arise from missing detections. This yields clear gains on ECP and eTRIMS, confirming the importance of explicit vertical modeling.

The full model (v3) further improves performance on Amsterdam, ECP and ParisArtDecoFacades, demonstrating the benefit of the floor-query assignment head on facades with regular vertical layouts. However, on eTRIMS, where unrectified viewpoints weaken vertical regularity, v3 underperforms v2, indicating that the assignment head may introduce noise when its structural assumptions are violated.

\subsection{Floor Assignment Analysis on Facades}
\label{sec:irregular_analysis}


We analyze the soft element-to-floor behavior of the assignment head using the GATA2Floor model trained on different datasets (ECP, eTRIMS, and ParisArtDecoFacades). For each detected window or door candidate (node), the assignment head produces a probability distribution over floor IDs and we pick the floor ID with the highest probability.

\begin{figure}[t]
    \centering
    \includegraphics[width=0.98\linewidth]{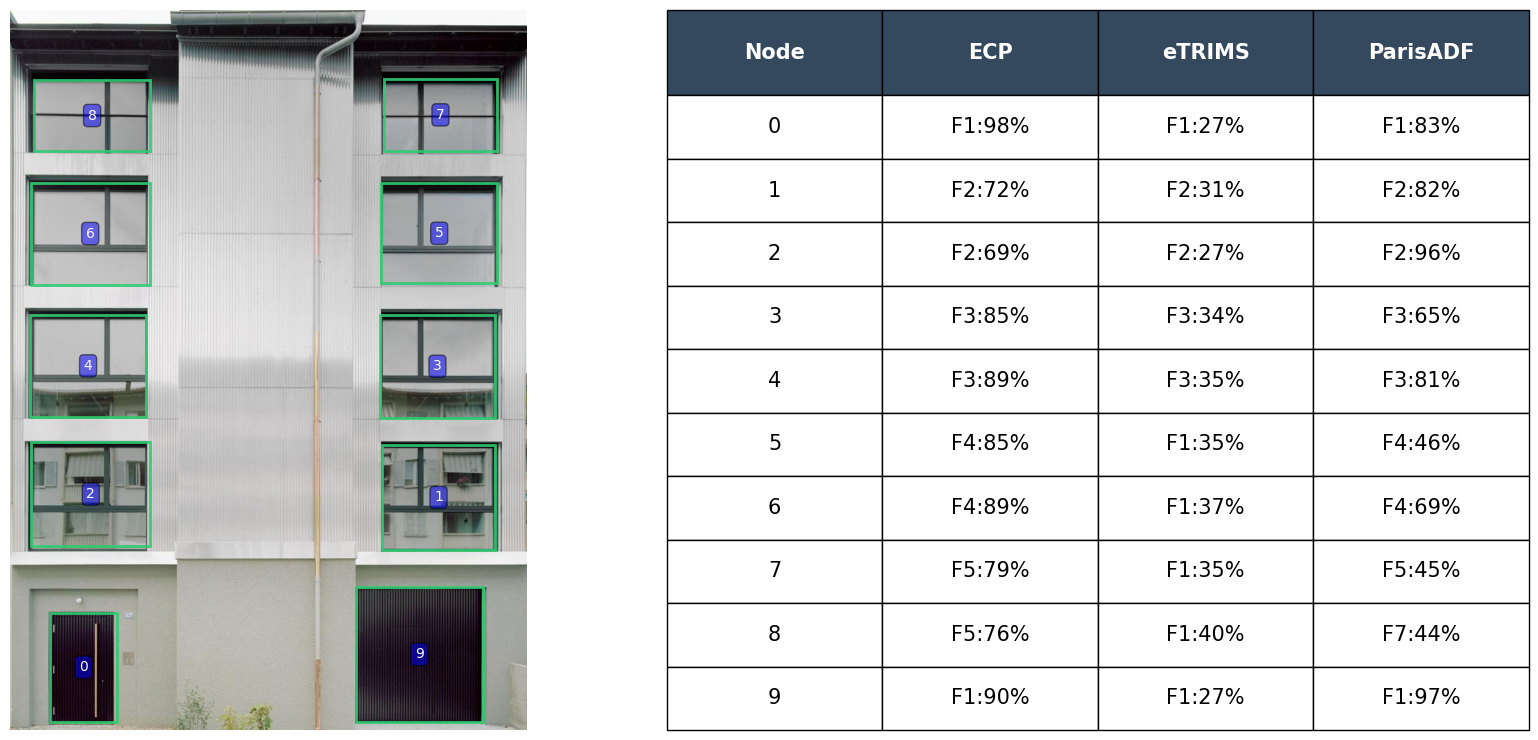}
    \caption{\textbf{Regular facade.} Left: detected facade elements with the graph node indices. Right: per-node floor assignments of GATA2Floor trained on ECP, eTRIMS, and ParisArtDecoFacades. Each table cell reports the highest predicted probability with F\# corresponding to the predicted floor ID.}
    \label{fig:regular_soft}
\end{figure}

\begin{figure}[t]
    \centering
    \includegraphics[width=0.98\linewidth]{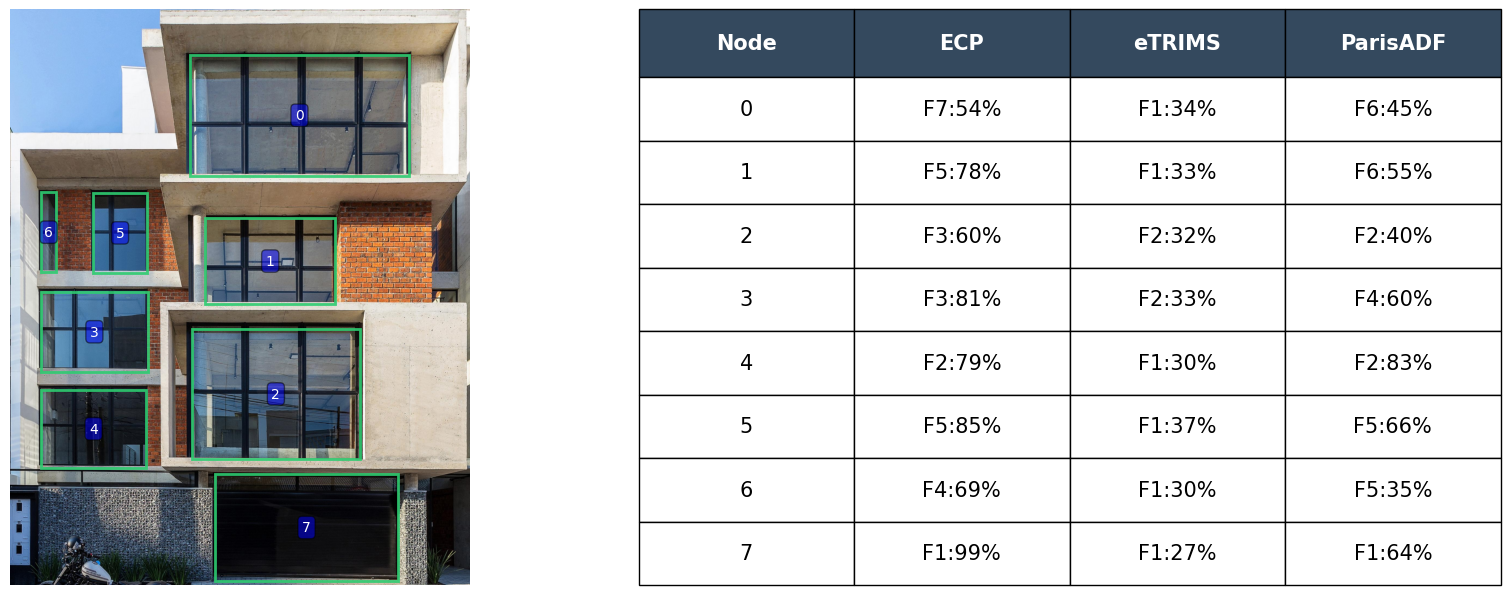}
    \caption{\textbf{Irregular facade.} Left: detected facade elements with the graph node indices. Right: per-node floor assignments of GATA2Floor trained on ECP, eTRIMS, and ParisArtDecoFacades. Each table cell reports the highest predicted probability with F\# corresponding to the predicted floor ID.}
    \label{fig:irregular_soft}
\end{figure}

As illustrated in Fig.~\ref{fig:regular_soft}, when the facade exhibits a consistent vertical arrangement, the dataset-specific models concentrate probability mass on a single floor for many nodes, producing stable per-node assignments. Conversely, Fig.~\ref{fig:irregular_soft} shows that on irregular structures, nodes often receive a spread-out floor distribution (e.g., mass split across nearby floors) and different models may favor different floors, providing an explicit view of uncertainty instead of only a single hard label.

Unlike black-box classification baselines (e.g., ResNet50), GATA2Floor enables inspection of \emph{which} elements are uncertain and \emph{why} assignments are ambiguous—windows at transitional heights between floors receive split probabilities, revealing architectural ambiguity rather than model error.

\subsection{Label-free proposal quality for graph construction}
\label{subsec:label_free}

We evaluate the label-free proposal mechanism as a fallback for graph construction when annotations are unavailable. We use negative prompting for CLIP~\cite{radford2021CLIP} and binary verification for GPT-4o, and include YOLO World~\cite{cheng2024yoloworld} as a generic open-vocabulary baseline. Proposal sources are compared based on their downstream effect on graph-based floor reasoning (Tab.~\ref{tab:label_free_downstream}). As shown, VLM-based proposals achieve substantially higher coverage than YOLO World, resulting in lower errors and non-zero tolerance accuracy, while YOLO World's near-zero coverage causes frequent floor-level misses and downstream counting failure. These results show that under label-free operation, proposal coverage is the primary limiting factor rather than localization precision. While performance remains below the supervised-detector setting, this pipeline offers a practical fallback for cities lacking datasets with facade-element annotations.


\begin{table}[t]
\centering
\footnotesize
\caption{Effect of label-free proposal sources on downstream floor counting. Coverage rate measures the fraction of ground-truth floors with at least one proposal.}
\begin{tabular}{l|ccc}
\hline
\textbf{Dataset} & \textbf{GPT-4o} & \textbf{CLIP} & \textbf{YOLOWorld} \\
\hline
\multicolumn{4}{c}{\textbf{Mean Absolute Error (MAE)} $\downarrow$} \\
\hline
ECP & \textbf{3.04} & \underline{3.60} & 4.81 \\
ParisADF & \textbf{4.81} & \underline{5.30} & 6.25 \\
\hline
\multicolumn{4}{c}{\textbf{Off-by-1 Accuracy} $\uparrow$} \\
\hline
ECP & \textbf{0.14} & \underline{0.09} & 0.05 \\
ParisADF & \textbf{0.12} & \underline{0.06} & 0.00 \\
\hline
\multicolumn{4}{c}{\textbf{Coverage Rate} $\uparrow$} \\
\hline
ECP & \underline{0.57} & \textbf{0.63} & 0.23 \\
ParisADF & \underline{0.47} & \textbf{0.53} & 0.17 \\
\hline
\end{tabular}
\label{tab:label_free_downstream}
\end{table}

\section{Conclusion}
\label{sec:conclusion}
This work models facades as vertical-aware graphs over window/door detections and introduces GATA2Floor, a multi-head GATv2 architecture that jointly performs global floor counting and soft element-to-floor assignment. Extensive experiments across public and a large unlabeled datasets show that GATA2Floor outperforms clustering-based baselines on floor counting, while a lightweight label-free proposal stage sustains performance when supervised detectors are unavailable. Together, these results demonstrate that relational reasoning with explicit vertical priors provides a robust inductive bias for facade structure, even under viewpoint changes, occlusions, and irregular layouts. 

\bibliographystyle{IEEEbib}
\bibliography{strings,refs}

\end{document}